# Shared perception is different from individual perception: a new look on context dependency

Carlo Mazzola, *Member, IEEE*, Francesco Rea, *Member, IEEE* and Alessandra Sciutti, *Member, IEEE*

*Abstract*— Human perception is based on unconscious inference, where sensory input integrates with prior information. This phenomenon, known as context dependency, helps in facing the uncertainty of the external world with predictions built upon previous experience. On the other hand, human perceptual processes are inherently shaped by social interactions. However, how the mechanisms of context dependency are affected is to date unknown. If using previous experience – priors – is beneficial in individual settings, it could represent a problem in social scenarios where other agents might not have the same priors, causing a perceptual misalignment on the shared environment. The present study addresses this question. We studied context dependency in an interactive setting with a humanoid robot iCub that acted as a stimuli demonstrator. Participants reproduced the lengths shown by the robot in two conditions: one with iCub behaving socially and another with iCub acting as a mechanical arm. The different behavior of the robot significantly affected the use of prior in perception. Moreover, the social robot positively impacted perceptual performances by enhancing accuracy and reducing participants' overall perceptual errors. Finally, the observed phenomenon has been modelled following a Bayesian approach to deepen and explore a new concept of shared perception.

## I. INTRODUCTION

Human perception integrates sensory information and predictions about the external world, a phenomenon that Helmholtz described in terms of unconscious inference [1]. Thus, sensory inputs are influenced by the previous experience organized along internal models acting as priors. A large body of research established that these two sources of information are integrated through Bayesian principles in many tasks, such as perception of an object (for a review see [2]), visual speed [3], [4], time intervals [5]–[8], categories [9], lengths [10], [11] and spatial localization [12]. The use of priors improves the reliability of perception, reducing the overall noise and is often considered to reflect a statistically optimal computation [13]. The influence of priors increases in the presence of low reliability of sensory input to cope with the uncertainty of the external world. For instance, this happens when the noise is due to an increased vagueness of the sensory information [12] or, on the other side, in people with lower perceptual acuity [6].

The phenomenon of context dependency, which had already been described by [14] with the name of central tendency, has been modelled in terms of Bayesian prior integration [5]–[8], [10]. When exposed to a series of stimuli of the same type, the perception of one stimulus is affected by the stimuli perceived before, so that their reproduction tends to gravitate toward their arithmetic mean. Therefore, perception is affected by the previous experience, built throughout the exposition of the entire series of stimuli. Such experience acts as an internal predictive model, a prior, on the incoming stimuli to reduce the variability of responses. Hence, prior influence produces an increased precision, even if at the expense of accuracy, resulting in a minor perceptual error as an overall effect. However, this beneficial effect on individual perception could hinder the efficacy of an interaction. Relying on previous personal experience, rather than trying to maximize the perception of the current stimulus, could cause misalignment with another agent having a different prior history, preventing effective coordination.

On the other hand, social interactions require establishing a common ground with the partner [15]. Without it, interactants would make nonsense of any verbal or non-verbal communication, causing misunderstandings, ambiguities, lack of coordination or perceptual mistakes. Even though different people might experience different perceptions of the same environment – opposite perspectives or the most varied emotional states – they commonly succeed in interacting with others by bridging these differences. How is this achieved when the difference between two individuals' perception stems from different prior histories?

In this work, we address the question of the role of internal predictive models on perception in a shared environment. Do humans maximize individual perceptual stability using internal priors, or do they align perception with the partner to facilitate coordination by limiting the reliance on individual priors?

Social interactions shape several human perceptual and cognitive processes. From the age of 1-year, selective attention is influenced by the direction of the partner's gaze [16], being at the basis of an ontogenetic process that will lead to other

This work has been supported by a Starting Grant from the European Research Council (ERC) under the European Union's Horizon 2020 research and innovation programme. G.A. No 804388, wHiSPER

C. M. Author is with the RBCS Unit, Istituto Italiano di Tecnologia and with the DIBRIS, Università degli Studi di Genova, Genoa, Italy (e-mail: carlo.mazzola@iit.it).

F. R. Author is with the CONTACT Unit, Istituto Italiano di Tecnologia, Genoa, Italy (e-mail: francesco.rea@iit.it).

A. S. Author is with the CONTACT Unit, Istituto Italiano di Tecnologia, Genoa, Italy (e-mail: alessandra.sciutti@iit.it).

interactive behaviors [17]. For instance, the ability to take the perspective of another person [18]–[20] seems to have its origins in this developmental process [21]. Furthermore, sociality impacts gaze movements [22], memory processes and information encoding at different levels [22]–[25]; it affects the processes of perception-action underlying joint-action (e.g. see the Joint Simon Effect [26], [27]), and influences the perception of space [11], [28], [29]. Therefore, we believe that a social context could significantly shape also basic perceptual mechanisms, such as context dependency.

To address this kind of questions, which explore the concept of shared perception, it is necessary to move the investigation from an individual, passive approach to an interactive shared context. To this aim, we propose to employ a humanoid robot as an experimental tool to investigate how perceptual mechanisms change during interaction. Cognitive science research studying the influence of a social context on perception may indeed benefit from the use of embodied artificial agents such as robots [30]. On the one hand, such complex sensory-motor devices allow generating controlled and precise actions in a repeatable manner. That enables the experimenter to replicate the rigorous control on stimuli traditionally adopted in the standard perceptual investigations within an interactive setting. This approach grants a degree of reproducibility of the (social and non-social) stimuli, which human actors cannot guarantee. On the other hand, robots ensure the experimental set-up an ecological layout given by their embodied presence in the shared physical space, instead of the virtual presence of an agent shown on a screen.

Extensive evidence shows the feasibility of the approach, demonstrating that robotic platforms can evoke social effects on humans, similar to those observed in human-human interactions. For instance, a robot can establish joint attention with users and elicit inferences about the intended referent [31]. Its behavior induces the same brain processes as if it was provoked by a human agent [32]. It has also been shown that robots elicit the same cognitive mechanisms of visual perspective-taking (VPT) that usually are elicited by human agents. The human partner spontaneously takes the visual perspective of the robot on a shared target, primarily when the robot directs its gaze or performs a reaching action toward it [33]. Moreover, Joint Simon Effect has also been found during interaction with robots [34], [35], suggesting that humans implicitly represent robots' actions as other humans' ones during joint actions.

In the current work, we will investigate the impact of social interaction on the perceptual processes of prior integration. We will ask participants to perform a perceptual task – estimating stimulus length – in a social and non-social scenario and assess whether perceptual performances change and whether they follow the prediction of a Bayesian model of context dependency. We will use a humanoid robot as a stimulus demonstrator to keep the same stimulation and just manipulate the context to make it either social or non-social. Previous results of our group [11] suggest that a robot exhibiting a social behavior can change the influence of priors on visual perception of space. This work will attempt to provide an in-depth analysis of the implication of such changes on perceptual performances and discuss how the Bayesian modelling approach adopted for modelling individual perception does not account for perception during interactions.

## II. MATERIALS AND METHODS

The present research was conducted to understand whether, during social interaction, human perception complies with the same principles of optimization it follows in individual scenarios [6], [10]. To this aim, we designed a user study to explore how the perceptual phenomenon of context dependency is affected by interaction with a humanoid robot acting as a mechanical or social agent, depending on the experimental session. The present work builds upon the methodology of two previous studies conducted by our team [10-11]. In [11], we approached the research by focusing on the regression index. Here, we deepened the analysis of the perceptual data to understand the impact of robot's behaviour on perceptual errors and to model the data in a Bayesian fashion [10]. Hence, for the reader, it is possible to refer to these former studies for further details of the methodology.

### A. Participants' demographics and Ethics

The experiment involved 30 participants (15 F, 15 M) over the age range of 19-46 years ($M$=28, $SD$=6). 37 % of them had already been exposed to interaction with the robot employed for the research (iCub). Nobody was aware of the purpose of the study. Due to technical problems, 3 participants could not finish the experiment, whereas other 2 participants had been excluded as outliers (see par. II. E. 3) so that in the end the sample was composed of 25 participants (13 F, 12 M). All of them signed a written informed consent before the experiment and received an honorarium previously agreed of 15€ for their time. The research had been approved by the regional ethical committee (Comitato Etico Regione Liguria).

### B. Procedure

The whole study involved four different counterbalanced within-subject conditions. In two of them, the participants interacted with the robot, whereas the other two tasks were performed individually.

*1) Set-up*

Participants performed all the tasks in the same experimental room where they sat at 50 cm from a touchscreen placed on a base 75 cm tall. In the task performed with the robot, the robot was placed on a fixed platform at 20 cm on the other side of the touchscreen, whereas during the individual sessions, it was hidden behind a curtain. Fig. 1a reports a schema of the experimental room.

In this study, we assessed how prior influence is altered when perceiving stimuli provided by another agent. To this aim, we needed an agent who acted as stimuli demonstrator reliably and consistently with all participants. We thereby employed the humanoid robot iCub [36], which is capable both to show a social behavior and to generate controlled and precise actions to replicate the rigorous protocol adopted in standard perceptual studies. The behavior of iCub was controlled to perform humanlike minimum jerk movements with an average hand speed of about 0.1 m/s. Specifically, the robot iCub presented the stimuli to participants by touching the screen and moving its torso and right arm according to models of biological motion.

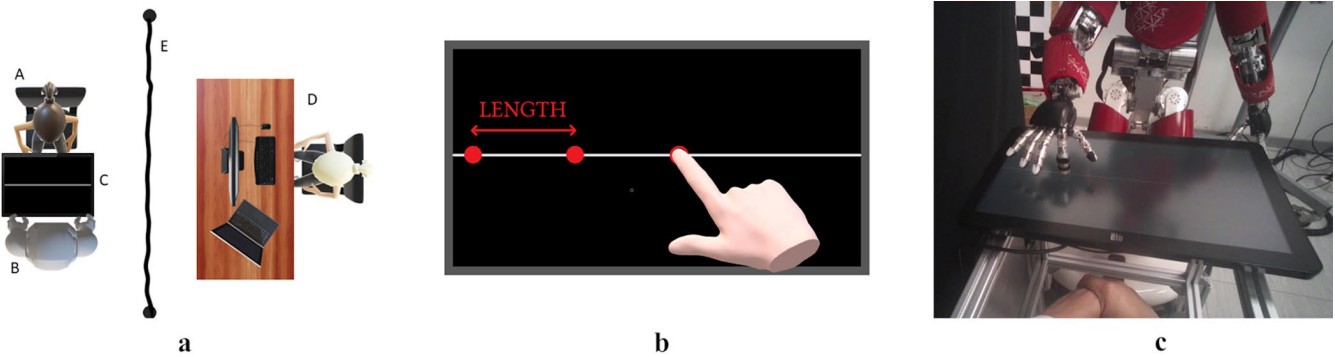

**Fig. 1.** Experimental set-up. Fig. 1a. Setting of experimental room: (A) participant's place, (B) iCub robot's place, (C) Touchscreen, (D) experimenter's desk, (E) curtain to divide and hide the experimenter's desk from participant's view. Fig. 1b. Description of Individual length reproduction task. Two dots are presented consecutively on a white line on a touchscreen, showing a certain length. Participants had to keep the second dot as a reference and to touch the screen in a third point, to reproduce the length of the stimulus. Fig. 1c. iCub from participants' perspective while touching the screen to present stimuli: images were obtained from Tobii Pro Glasses 2 recording.

A widescreen LCD Touchscreen Monitor ELO 2002L 20-inch was employed with a resolution of 1920x1080px for an active area of 436.9mm x 240.7mm, at a frequency of 60Hz and Response Time of 0.02 s. The monitor was positioned horizontally: it showed the stimuli to participants and recorded both the touches of the robot and the responses of participants. It was programmed with MATLAB 2019a with Psychophysics Toolbox Version 3 (PTB-3) and controlled by a Windows 10 pc. To record participants' gaze information during the interaction with iCub, we asked them to wear a Tobii Pro Glasses 2 (100 Hz gaze sampling frequency).

*2) Experimental Sessions*

To test the experimental hypotheses, we set up different sessions. An individual task of length reproduction served as a baseline to assess participants' level of context dependency. The other two sessions were performed with the robot acting differently, as a mechanical or social agent, to determine how social interaction affects context dependency in perception.

*a) Individual length reproduction task*

In the individual length reproduction session, the participants' task was to reproduce the lengths indicated by two dots presented on the screen by touching the screen in a third point (see fig. 1b). Specifically, the reproduced distance – between the second dot and the point touched by participants – should be equal to the presented length. The stimuli were presented as two consecutive red circles of 1 cm diameter lasting 0.6 s each and appearing with an interval of 2 s. The first dot was presented at a variable distance from the left border of the screen (0.5–3.5 cm, randomly selected). The second dot was shown at the right of the first one, at a distance of 11 different lengths from 6 cm to 14 cm with a difference of 0.8 cm each. Each distance was presented 6 times, randomly, for a total of 66 trials with additional 3 practice trials. After the response, another equal red disk appeared in the touched point, but no feedback was provided about the accuracy of the response.

*b) Length reproduction tasks with the robot*

In the two main sessions of the experiment, participants interacted with the humanoid robot iCub. iCub acted as a stimulus demonstrator touching the screen in the two endpoints of the lengths (see fig. 1c). Participants' task was the same as in the individual length reproduction task (see fig. 2a). The touchscreen did not show any light in the points where iCub or the participants touched. The robot was programmed to present the same stimuli of the individual task. Whereas participants' task was the same in both conditions, the behavior of the robot changed from one condition to the other. Indeed, for a correct evaluation of the impact of sociality on perception, it was needed to compare two conditions where the very same sensory inputs were presented as stimuli, and only the nature of the presenter (social vs mechanical) was manipulated. In this case, the stimuli were always provided by the robot's finger indicating two points on the touchscreen, with the very same kinematics in two different conditions, "Social" and "Mechanical". For other specifics about the procedure see [11].

*C. Characterization of robot's behaviour*

Since the research aimed to study the perceptual alteration induced by sociality with the aid of a robotic stimuli demonstrator, we decided to differentiate as much as possible the participants' perception of the robot in the two conditions. Implicit behavioral and verbal cues of the robot were therefore combined with explicit priming of participants about the robot's intentionality and skills.

In the social condition with the robot, iCub acted as an interactive social agent. Its left eye camera was turned on to track participants' faces and establish mutual gaze before starting the task, after its end, and between one trial and another, to give an implicit idea of turn-taking (see fig. 2a-b). Moreover, iCub showed emotions with its facial LEDs: it mostly smiled with a friendly expression, unless while touching the screen, when it was programmed to appear focused on the task. Through the iKinGazeCtrl Module [37], iCub also exhibited natural oculomotor coordination with its hand by directing its gaze in advance toward the point it was going to touch. Before starting the practice trials, iCub welcomed participants and explained to them the task: "*Hi, I'm iCub! Now, we will play together. I will touch the screen twice, and you will touch the screen a third time to replicate the distance. Are you ready?*". Then after 33% and 66% of trials, the robot incited participants with these words: "*Well done! Keep it up!*" and "*Come on, there are only a few more trials left, keep focused*". Finally, at the end of the task: "*Thank you for having played with me! It has taken a bit of a long time, but you are helping us a lot! See you next time*".

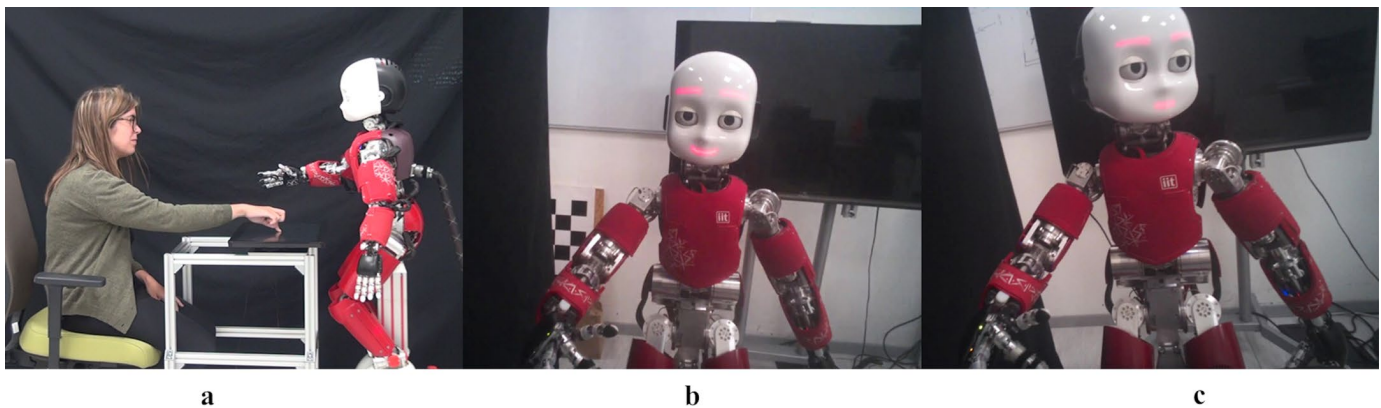

**Fig. 2**. The interaction with iCub. Fig. 2a. Picture of the interaction with the robot in the reproduction task (social condition). Fig. 2b-c. Participants' perspective on iCub during the reproduction task. Pictures were obtained from the recordings of the Tobii pro-glasses 2 respectively during the interaction with the social robot while exchanging mutual gaze with it (fig. 2b) and with the mechanical robot, whose head position was fixed with the face turned away from the participants (fig. 2c).

On the other hand, in the mechanical condition, iCub acted as a mechanical agent without showing any social feature. To this aim, iCub head joints were fixed so that its head was turned away from the participants (see fig. 2c). This behavior was designed to show that the robot had no awareness of the environment or the task. Also, face-LEDs were static, so that it appeared without emotions, and the robot did not talk. The only parts that were moved were the joints of the torso and the right arm, like a robotic arm.

To strengthen the differentiation of the two conditions, the experimenter diversified the introductory explanation of the task when talking about the robot. Outside the experimental room, in the social condition, the researcher introduced the session in this way: "*Now iCub is fully working, with its social intelligence on and its cameras are switched on to look at you and the screen. It will be showing you two positions on the touch screen. Please reproduce the distance between these two points by pressing the touchscreen in a third one at equal distance from the last shown by the robot*". Conversely, before starting the task, participants were instructed with these words by the experimenter: "*In this session, iCub's social intelligence is turned off. The computer is just driving its hand motions in a predefined pattern. It will be touching two positions on the touch screen. Please, reproduce the distance between these two points by pressing the touchscreen in a third one at equal distance from the last one*".

*D. Questionnaires*

Besides a set of questionnaires delivered to participants before the experiment to understand their demographic, their personality and attitudes (TIPI-test [38] and AQ test [39], [40]) and their attitudes toward robots NARS questionnaire [41]), another set of questionnaires was submitted after each session with the robot to check the manipulation effect of the robot's behavior and explicit priming. To this aim, once participants ended each task with the robot, they were asked to go out of the room and fill a form of questions online. On this occasion, we delivered the Inclusion of Other in Self-scale (IOS) questionnaire [42] to assess how close to iCub, in a range from 1 to 7, participants felt during the task; the Godspeed questionnaires with the sub-scales Anthropomorphism, Animacy, Likeability and Perceived Intelligence [43] and the subscales Mind experience and Mind agency of a Mind perception test [44], [45]. We proposed all of them on a 7-points Likert scale. At the end of the experiment, a final questionnaire for debriefing was provided to participants to understand their opinions and feedbacks about the tasks and the behavior of iCub.

*E. Data Analysis*

*1) Length Reproduction*

To investigate the phenomenon of context dependency, we analyzed the reproduced lengths following a well-established approach [6], [10]. The influence of prior experience on sensory stimuli, which occurs as the integration of different kinds of information, can also be interpreted as the dependence of perception on its context. For instance, in visual perception of space, perception of a visual stimulus is affected by distances experienced before, which cause a perceptual bias. The overall effect resulting from such integration is thereby a regression of all perceived stimuli toward the mean of the presented stimuli, which act as prior built during the exposition to all the set of stimuli. In this way, the long distances are perceived as shorter than they are and vice versa. Regression Index is, therefore, a direct measure of the degree of context dependency, which is computed as the difference in slope between the identity line (stimuli-correct responses) and the best linear fit of the reproduced values plotted against the related stimulus (see fig. 3). The index varies from 0 (no regression) to 1 (complete regression). Specifically, in our study, the stimuli were presented to participants so that their arithmetic mean was 10 cm.

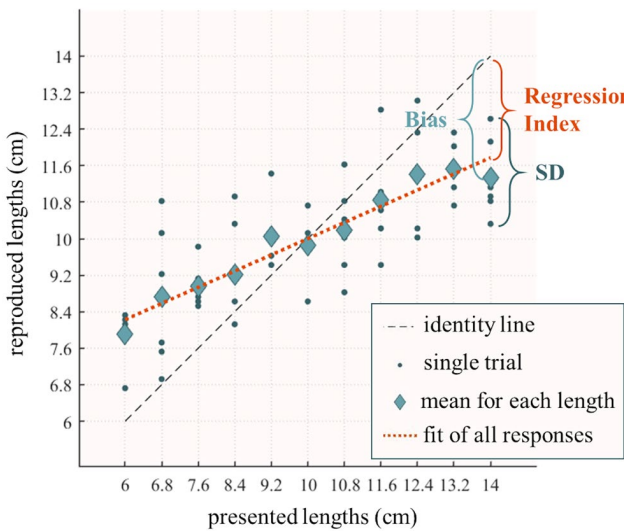

**Fig. 3.** *Plot of the data of a length reproduction task.* Reproduced lengths are plotted against the related stimuli. The regression index is calculated as the difference between the slope of the linear fit of the ideal reproductions (identity line) and the slope of the linear fit of the real data. For each stimulus we measured the average bias and the Coefficient of Variation of the related responses.

Since the present research aimed mainly to assess and model the perceptual errors associated with the phenomenon of context dependency, we first portioned the total error of responses in two parts: the bias and the coefficient of variation (CV) that respectively measure participants' accuracy and precision. To calculate these two parameters for each participant, a constant bias has been removed from each response by subtracting the average response of all trials and summing the length of the average stimulus.

Then, for each *i*-th stimulus, we measured bias as the difference between the average response for that stimulus ($R_{Mi}$) and the stimulus ($S_M$), in absolute value, normalized for the average stimulus of the entire session ($\bar{S}$). In the robot sessions, since motor noise caused a slight imprecision in the stimuli demonstration, we used the average stimulus presented by iCub for each of the 11 lengths ($S_{Mi}$).

$$BIAS_i = \frac{|R_{Mi} - S_{Mi}|}{\bar{S}} \quad (1)$$

On the other hand, the CV of responses of each stimulus was calculated from the standard deviation of the responses to that stimulus, again normalized for the average stimulus of the entire session ($\bar{S}$).

$$CV_i = \frac{\sqrt{\frac{\sum (R'_i - \overline{R'}_i)^2}{N}}}{\bar{S}} \quad (2)$$

Finally, the normalized total error is calculated for each stimulus as the root-mean-square error (RMSE) from the bias and the CV.

$$RMSE_i = \sqrt{BIAS_i^2 + CV_i^2} \quad (3)$$

Statistical analyses of the data related to perceptual errors in the three conditions were conducted using the Linear Mixed Models in R with specific libraries [46], [47].

*2) Gaze analysis*

To assess possible variations in the way participants visually interacted with the robot in the two conditions, we analyzed data of participants' gaze gathered through a gaze-tracker, the Tobii Pro Glasses 2, during the task performed with iCub. This information also served as an additional behavioral check of the manipulation of iCub's social features to understand whether the robot was also recognized implicitly by participants. The number of times participants looked at iCub's face during the experiment served to measure participants' involvement during the interaction. To extract these data, we firstly obtained the images of the iCub's face from Tobii recordings. We, therefore, trained the software of the gaze-tracker (Tobii Pro Lab) to recognize the face of iCub as a region of interest in the recordings to check whether participants' looks stopped on the robot's face. In this way, we counted the percentage of times in which participants looked at iCub's face during each session. We assessed such a percentage by counting the number of trials in which iCub's face was looked at least one time. Specifically, the measure was taken for two kinds of time intervals: the interval between the first and the second touch of iCub (during trials) and the interval between the second touch of iCub and the first one of the subsequent trial (between trials) (See fig. 4).

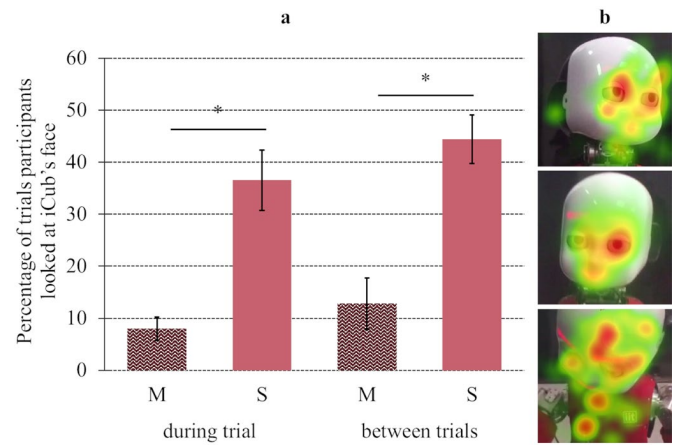

**Fig. 4.** Participants gaze towards iCub face. Fig. 4a. Bar plot of the % of trials in which participants looked at iCub face during trials (tot trials = 66 trials) and between one trial and another (tot intervals = 65) in the mechanical and in the social condition. Fig. 4b. Heatmaps of participants gaze on three representative snapshots referred to the mechanical condition (the one above) and to the social condition (the two below)

We conducted statistical analyses to compare participants' gaze data in the two tasks with the robot with Jamovi 1.6.1 [48]. Data have been extracted using Tobii Pro Lab Software and Python with Pandas Data Analysis Library. Due to technical problems with the device and because some of the participants wore their glasses, we could analyze only 15 participants from our sample.

*3) Perceptual ability check and outliers*

We organized a perceptual task of length discrimination to assess whether participants were able to perceive the visual stimuli reliably or whether all their performances should be discarded (See [11] for details about this task). Specifically, we decided not to analyze participants who revealed not being able to discriminate a distance smaller than 4 cm, which is the difference between the mean stimulus of the reproduction task and the extreme ones.

We also decided to exclude participants whose performance in the reproduction tasks exceeded the average performance of all participants of at least 2.5 times the SD of the sample. We removed two participants from the sample after this last screening.

*F. Bayesian Modelling*

Context dependency is a perceptual phenomenon that can be explained as integration between sensory information (each current stimulus) and priors (built on the stimuli already perceived). Previous research has demonstrated that such phenomenon can be described in a Bayesian fashion and follow Bayesian principles of optimality [5]–[7], [10]. Specifically, although leading to inferior accuracy in the outcome of the perceptual process, the influence of priors enhances precision – and the overall total error – by reducing the variability of the responses.

The present research aims therefore to analyse the influence of priors on visual perception of space by connecting with previous studies and, for the first time, to assess the effect of sociality on context dependency with a Bayesian approach.

In this perspective, following the approach proposed by [6], the perceived length of a stimulus (Posterior) can be modelled as a gaussian defined by $\mu_R$ and $\sigma_R$, and resulting from the product of other two Gaussians (see Fig. 5): the current noisy sensation of the stimulus length, represented by the Likelihood, and the Prior, which is an estimate of the series of stimuli previously perceived.

For each stimulus, the Likelihood function is modelled as a Gaussian centered on the actual length of the stimulus ($\mu_L$) with standard deviation ($\sigma_L$) corresponding to the sensory precision of each participant. Conversely, the Prior is modelled as a gaussian distribution with the mean ($\mu_P$), corresponding to the average stimulus of the series, and an amplitude ($\sigma_P$) that represents the weight given to the prior during perception. Thus, according to the model, given a fixed prior width, the observers' response is derived as a function of their sensory precision: the better it is (i.e., the narrower the likelihood distribution is), the nearer the response will be to the sensory information. On the contrary, the worst observers' sensory estimate is, the closer their response will move towards the prior.

Given these premises and according to Bayes' rule, it is possible to calculate the mean and the standard deviation of the posterior distribution (for formulas see [10]). Accordingly, both the bias and the variance of the observer are computable, and from (3) also the RMSE.

In this way, from the data obtained in the three reproduction tasks, it has been possible to model the perception of participants, compare our results with the previsions of the model, and understand how social interaction impacts on the use of prior knowledge.

The analyses and the simulation of the Bayesian Model were conducted with MATLAB 2020A.

## III. RESULTS

Our research was founded on the primary hypothesis that interaction with a social agent plays a role in how humans perceive space. We aimed to assess whether interactive scenarios impact human integration of visual information with prior and, if it happens, how error parameters of perception, namely, accuracy and precision, are affected.

*A. Manipulation Check*

From the questionnaires completed after each interaction with the robot, we could verify whether iCub's behaviors in the "mechanical" and "social" conditions were effectively perceived as significantly different. Table 1 reports all the scores of the scales provided in the questionnaires. When iCub behaved socially, it was perceived as significantly more anthropomorphic, animate, intelligent and likeable. In addition, in that condition, participants attributed more extensively to him a mind and an experience. Finally, they also felt closer to him.

Behavioral measures of gaze collected through the Tobii Pro Glasses 2 confirmed that participants recognized the diverse behavior of iCub also implicitly, not only when asked through questionnaires. They looked indeed at the face of the robot significantly more often in the social condition than in the

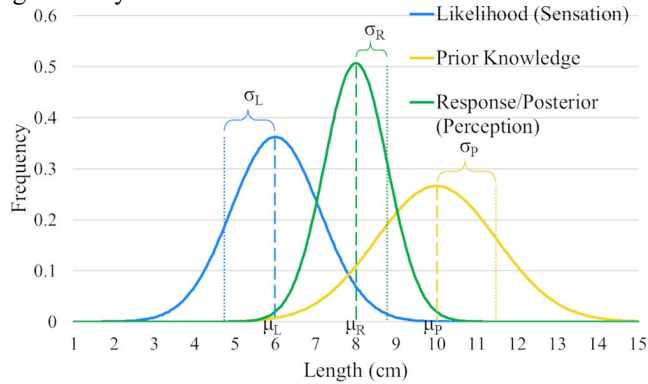

**Fig. 5.** Representation of Bayesian Model (modified by [6], [10]). Perception (Posterior distribution) is described as a Gaussian resulting from the integration between the Likelihood distribution of the stimulus of length $\mu_L$ with sensory precision $\sigma_L$ and the Prior distribution centered in $\mu_P$ with a weight of $\sigma_P$.

TABLE I
MANIPULATION CHECK WITH DATA ACQUIRED FROM THE QUESTIONNAIRES PROVIDED AFTER EACH TASK WITH THE ROBOT.

| Feature | Min-Max Score | Mechanical | Social | Test |
|---|---|---|---|---|
| Anthropomorphism | 5-35 | *M*=12.8, *SD*=4.81 | *M*=19.2, *SD*=5.97 | W. S-R test Z=-3.78, *p*<.001 |
| Animacy | 5-35 | *M*=13.0, *SD*=5.41 | *M*=21.7, *SD*=5.31 | W. S-R. test Z=-4.38, *p*<.001 |
| Likeability | 5-35 | *M*=22.0, *SD*=7.18 | *M*=28.8, *SD*=21.6 | W. S-R test Z=-3.72, *p*<.001 |
| Perceived Intelligence | 5-35 | *M*=21.6, *SD*=4.98 | *M*=24.6, *SD*=3.79 | W. S-R test Z=-3.18, *p*<.005 |
| Mind experience | 4-28 | *M*=7.04, *SD*=4.74 | *M*=12.4, *SD*=7.66 | W. S-R test Z=-3.55, *p*<.001 |
| Mind Agency | 4-28 | *M*=11.9, *SD*=5.49 | *M*=16.7, *SD*=6.97 | W. S-R test Z=-3.57, *p*<.001 |
| Inclusion of other in the self-scale | 1-7 | *M*=2.84, *SD*=1.52 | *M*=4.44, *SD*=1.42 | W. S-R test Z=-4.09, *p*<.001 |

In column "Test": "P.S. T-Test" is paired sample t-test, while for not normal distributions of data "W. S-R Test" is the Wilcoxon Signed-Rank test.

mechanical one, both during trials, i.e. in the time interval between the first and the second touch of iCub (about 36% vs 8%, Wilcoxon Signed-rank test: Z=120, *p*<0.001), and between trials, that is in the time interval between the second touch of iCub and the first one of the subsequent trial (about 44% vs 13%, Wilcoxon Signed-Rank test Z=117, p=0.001).

### B. Context dependency and normalized errors

The main goal of this study was to understand the implication of a social scenario towards the use of priors in perception and to attempt a description of it using the Bayesian Model that up to now has been employed to describe perception only in individual scenarios [5], [6], [10]. Participants exhibited a significant context dependency (or regression to the mean) in the individual condition, with an average regression index of 0.446 (*SD*=0.133), significantly larger than 0 (one-sample t-test, t(24)=16.8, p<0.001, Cohen's d=3.36). Participants' perception was influenced by prior knowledge, leading to overestimating the shorter stimuli and underestimating the larger ones. In the two conditions with the robots, participants still show a context dependency phenomenon, although to a lesser degree (*M*=0.263, *SD*=0.175, one-sample t-test, t(49)=10.6, p<0.001, Cohen's d=1.50).

More interestingly, participants exhibited a significantly lower degree of context dependency in the social-robot condition than in the mechanical one (mech: *M*=0.292, *SD*=0.183; soc *M*=0.234, *SD*=0.165), notwithstanding the sensory stimuli to be reproduced in the two conditions were identical. Indeed, a paired t-test comparing the two robot conditions revealed a significantly lower regression index in the social one (t(24)=2.92, p=0.007, Cohen's d=0.584), (see fig. 6).

To deepen the understanding of the influence of social interaction on perception, we also analyzed the errors of reproductions, evaluating accuracy (bias), precision (CV), and total error (RMSE), as described in Methods-Data Analysis (see fig. 7 and 9a). We ran three Linear Mixed Effect Models, with the average error (bias, CV or RMSE) for each of the 11 stimuli as a dependent variable and the condition (Individual, Mechanical, Social) as a predictor. Furthermore, we applied random effects to the intercept at subject and stimulus levels. The random effect at the subject level has been applied to adjust for each subject's baseline level of error and model intra-subject correlation of repeated measurements. The random effect at the stimulus level served to model inter-stimulus variability in the error parameters. Random effects were submitted to the model in this order.

Firstly, we assessed the shift of both the sessions performed with the robot from the pure individual condition. We found a significant decrease of the bias both in the mechanical condition (Mechanical – Individual: B= -0.019, t=-3.74, p<0.001) and in the social one (Social – Individual: B= -0.033, t=-6.53, p<0.001). Such a difference could be partially attributed to the richer information of the stimulus in the conditions with the robot. Indeed, whereas in the individual condition stimuli were shown only by a dot appearing on the screen, in the robot condition the information about the distances was given by the hand and the finger of iCub touching the screen. With regards to the CV, it was not found any significant variation, neither with the mechanical robot (Mechanical – Individual: B=

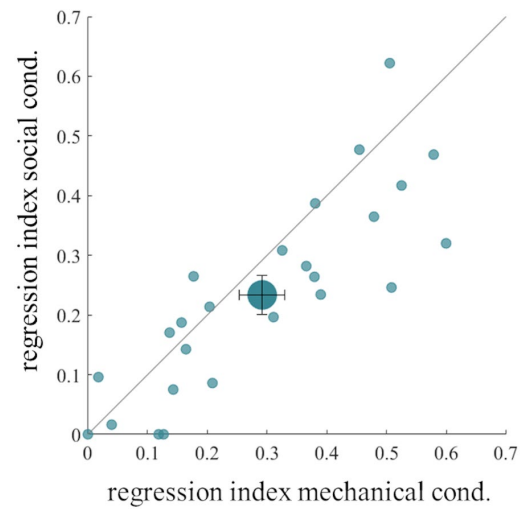

**Fig. 6.** Scatter plot of regression index values for each participant in the two conditions with the robot. The smaller dots represent single participants in the mechanical and the social condition, the largest one represents the mean with error bars calculated from the standard error of the two conditions.

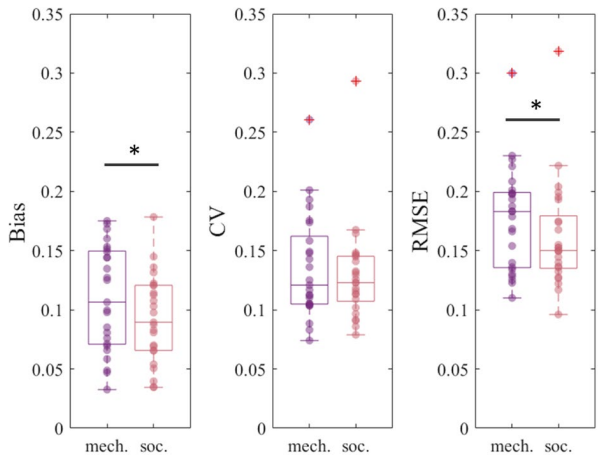

**Fig. 7.** Boxplot representing the values of perceptual errors (Bias, CV and RMSE) in the two conditions with the robot: mechanical and social. Perceptual errors have been normalized for the mean stimulus presented in the task (10 cm). Circles represent perceptual errors of each participant.

0.0005, t=0.818, p<0.414), nor with the social one (Social – Individual: B= -0.00006, t=-0.099, p<0.921). Conversely, the RMSE was found significantly lower in the social condition (Social – Individual: B= -0.007, t=-3.525, p<0.001), but not in the mechanical one (Mechanical – Individual: B= -0.022, t=-1.112, p=0.266)

Since the two robot conditions were more comparable in terms of richness of information of the stimuli presented by the robot, we focus more specifically on the difference between them to assess the variation caused by sociality. Thus, we directly compared the errors of the two sessions performed with the robot with the three Linear Mixed Effect Models. Results revealed a significant effect of the bias (Social – Mechanical: B=-0.014, t=-2.784, p=0.005) and of the RMSE (Social – Mechanical: B=-0.015, t=-2.407, p=0.016), which resulted lower in the social condition. No significant effect has been found for the CV (Social – Mechanical: B= -0.005, t=-0.911, p=0.362) (see fig. 7).

To further understand the variation of perceptual errors we observed between the two tasks with the robot, we also performed a statistical analysis to find whether it might be correlated with the variation participants revealed in how they perceived iCub's behaviour in the two conditions. For this analysis, we used both the data gathered from questionnaires and the behavioral gaze data. Results of a Spearman correlation indicated that a significant negative association was verified between the variation in bias (Δbias: social-mechanical) and the variation in the value of anthropomorphism (Δanthropomorphism: social-mechanical) ascribed to iCub in the two conditions: rs(25) = -0.498, *p*=0.011 (see fig. 8). Such a result revealed that the robot aspect was the most critical feature of iCub that had an impact on perceptual data. The reason is that the Anthropomorphism scale includes questions about participants' impressions of the robot in terms of being fake - natural, machinelike - humanlike, unconscious - conscious, artificial - lifelike and moving rigidly – elegantly (see [43]).

### C. Simulation of the Bayesian Model

In fig. 9a-b, our data are plotted within the Bayesian framework that models context dependency as described in Methods. The symbols in fig. 9a correspond to single participants' and average CV as a function of the corresponding Bias in the three conditions.

In terms of CV (precision), no difference is visible among the three conditions. On the contrary, considering accuracy, the bias of the three conditions decreases with this order: individual-mechanical-social condition. A similar pattern can be identified for the total error (RMSE), which can be seen as the distance from the axes-origin. The plot then clearly illustrates the results of the statistical analysis.

Starting from for 4 fixed values of $\sigma_P$ (0.5 cm, 1.5 cm, 2.5 cm, and 3.5 cm) and from $\sigma_L$ varying between 0 and 0.6, the continuous lines in the graph represent the model predictions for Bias and CV derived as described in Methods Par. F - Bayesian Modelling and normalized for the average stimulus (10 cm). As in [10], a further constant of 1.2 cm representing the non-sensory motor noise was also added to the CV. As shown in fig. 9a, the results of all three conditions are predicted by the model with a $\sigma_P$ of about 1.5 cm.

In fig. 9b, the 4 sigmoid lines represent the model predictions about the relations between regression index and $\sigma_L$ for the same 4 fixed values of $\sigma_P$ used in fig, 9a. The background of fig. 9b is color-coded to represent the different values of RMSE predicted by the model. The model predicts the highest values of RMSE when $\sigma_L$ is low and regression index is high: basically, in the case of an ideal subject who would have excellent eyesight but nonetheless relies heavily on its previous experience. Then, RMSE grows again when $\sigma_L$ is high, but the observer does not regress enough to mitigate the error caused by the weak eyesight. The lines derived from the model lie in the minimum of the RMSE as evidence of optimality.

We can assess where our data would lie on the model by considering the average regression indexes measured in the three different conditions of the experiment as our ordinates. Assuming that the three conditions share the same prior width (1.5 cm, as derived in fig. 9a), the model would predict that the perception in the three sessions was characterized by different $\sigma_L$.

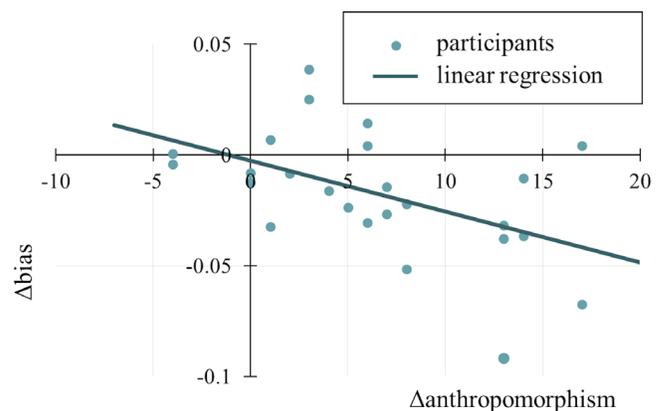

**Fig. 8.** Linear regression. For each participant, the variation of bias between the two conditions with the robot (social-mechanical) is plotted against the variation of the grade of anthropomorphism between the variation in bias (Δbias) and the variation of the grade of anthropomorphism (Δanthropomorphism) ascribed to iCub in the questionnaires after each condition (social-mechanical).

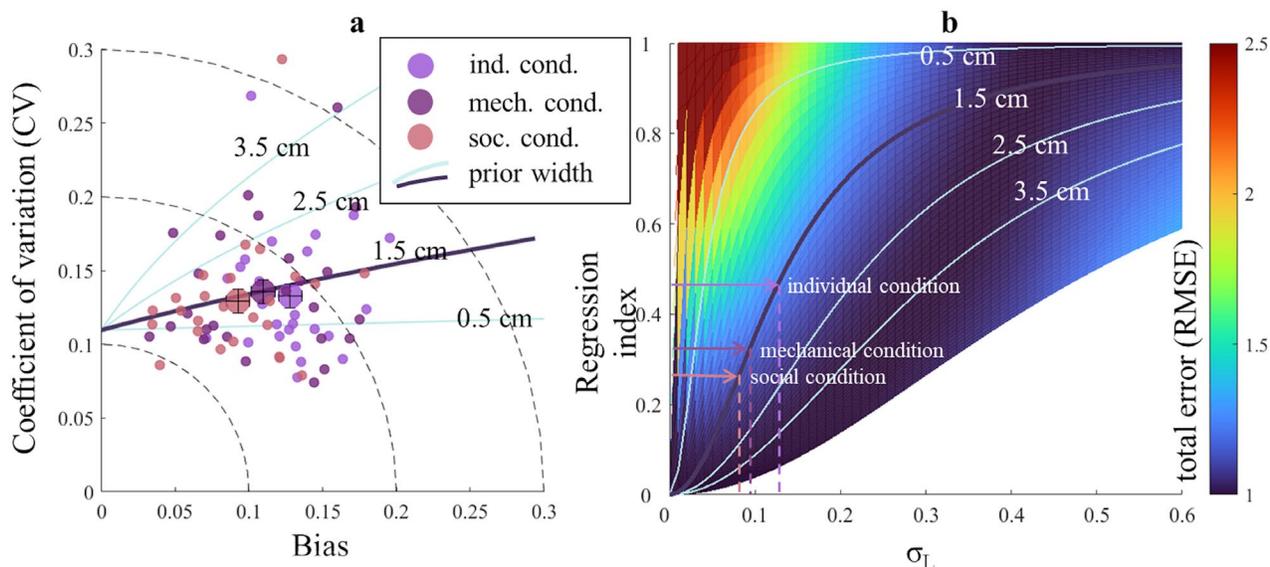

**Fig. 9.** Model simulation. Fig. 9a shows the portioned perceptual errors in the three conditions: large circles represent the average normalized CV plotted against the average normalized bias with the error bars representing the standard error; small circles are individual participants. The four curves represent the prediction of the Bayesian model given a fixed value of $\sigma_P$ (0.5 cm, 1.5 cm, 2.5 cm, 3.5 cm), which represents the weight given to the prior. Each curve has been plotted by varying $\sigma_L$ (Weber Fraction) from 0 to 0.6. As in [6] and in [10], an additive fixed non-sensory motor noise of 0.12 has been from [10]. In Fig. 9b, arrows represent the simulation of the model for $\sigma_L$, starting from the empirical data of the regression index and from the value of $\sigma_P$ derived by the model (Fig. A). In Fig. 9b it is also represented the value of RMSE simulated by the Bayesian model once given the regression index and $\sigma_L$ and normalized for the minimal values of RMSE related to each value of $\sigma_L$.

As we mentioned above, $\sigma_L$ is considered a function of the sensory threshold. Therefore, it depends on the observer's visual acuity or the richness of the stimuli's sensory information. A higher visual acuity – or more visible stimuli – correspond to lower $\sigma_L$. A difference in the nature of the stimuli is indeed present between the individual conditions and both the robot ones. With the robot, the stimuli were provided by a gesture of the humanoid, whereas in the individual condition, they were indicated only by the red dots appearing on the screen. The richer sensory information associated to the robot action might therefore explain the lower $\sigma_L$ in the robotic conditions. Conversely, between the Mechanical and the Social conditions there was no difference in the sensory information since the robot's movement was the same in both sessions. If we were to impose an equal $\sigma_L$ between the Mechanical and Social conditions – given that the participant's acuity does not change and neither the stimuli – the model would predict a lower prior weight (higher $\sigma_P$) for the social condition. However, this hypothesis would be incompatible with the measured CV and Bias in the social condition (see fig. 9a).

In summary, the switch from mechanically-generated stimuli to stimuli generated by a social agent – though physically identical – led to a different degree of context dependency in our participants. However, a model, which predicts the level of integration of prior experience in perception by uniquely basing the estimation on the sensory acuity of the observer or, in turn, on the physical properties of the stimulus that can affect its visibility cannot explain, alone, the data we collected.

IV. DISCUSSION

*A. Context dependency in social interactions*

As humans, we adopt effective strategies to reliably perceive what is around us, to interpret others' behaviour and, as a sum of the two things, to interact and coordinate with them in a shared environment. To do that, not only we consider the information coming from our senses, but we build and use internal models coming from previous experiences that help us to cope with the uncertainty of information. If we just remain at a purely perceptual level, perception can be seen as an inferential process where previous experience influences the percept by acting as prior toward the incoming sensory information. But how do we use such priors when interacting with another agent? Which influence do they have on our perception, for example, on our levels of accuracy and precision? And what such an influence can reveal about the way humans perceive and share the environment with others? The idea underlying this study aimed precisely to start answering these questions.

The Bayesian model defined in Methods has been so far used to study prior inference mechanisms in individual contexts of perception. No parameter is present to assess the variation that a social scenario could bring to perception. Therefore, it should be verified whether descriptive models of individual perception can account for the change induced by sociality and verified in the variation of perceptual errors. To achieve this, it has been used a humanoid robot as a reproducible and controllable stimuli demonstrator. This solution could combine the rigorous protocol adopted in standard perceptual studies with an embodied interactive context. First, our results indicate that the perceptual phenomenon of context dependency occurs even in a social-interactive context, where a social robot shows stimuli. This means that humans employ their priors even in a social interactive context to perceive the world around them.

The strategy of context dependency is put in place by our perceptual system to cope with the uncertainty of sensory input and to reduce variability at the expense of accuracy [5], [6], [10]. Thus, theoretically, the decrease in context dependency observed in the social robot condition could have been associated with increased responses variability. Conversely, our results show that the interaction with the social robot kept a positive impact on perception. With the social robot, participants demonstrated a significantly higher accuracy (lower bias) with respect to the interaction with the mechanical robot and to the individual condition, without having a negative influence on precision (CV). This implies that (1) participants were more focused on each current stimulus they received from the social robot, as revealed by their higher accuracy, and (2) they were not distracted by its social behaviour. Indeed, even their overall error in reproduction, as measured by RMSE, was significantly lower in the social condition.

Theoretically, in the Bayesian model, these results could be justified as a variation of two parameters: $\sigma_L$ and/or $\sigma_P$. Nevertheless, as resulted from the simulation of the model with our data, the shift between the social and the mechanical condition seems not to be explained either to an increased $\sigma_L$ due to more visible stimuli or higher sensory acuity, either to an increase in $\sigma_P$ that is a weaker prior. That being the case, the descriptive model based on individual perception does not account for the variation induced by the interaction with the social agent. Accordingly, in a more general model of context dependency, it is necessary to consider that the inferential processes of perception are a function of the social context, which could be described as reliance on one of the two sources of information: the current stimulus shared with the partner, or the private internal model built upon one's own experience. Thus, the perceptual mechanism of context dependency would depend also on the shared context of perception that may bring each partner to be more attentive towards the shared reality and to exploit less the private internal models about the world around.

This means that, given the exact same stimuli provided by the two robots, when interacting with the social robot, the inferential processes of perception are affected in favor of higher reliance on sensory information and a weaker dependence on the priors.

### B. Impact of robot's behaviour

The effect of robot sociality on human distraction has been studied in different tasks and is an open question in the field of human-robot interaction. On this issue, in [49] it has been evidenced that the social behaviour of the robot negatively affected the child learning with respect to mechanical behaviour. Authors hypothesized this could be due to distraction caused by the social robot or by a higher cognitive load induced by the social interaction. Furthermore, [50] found that in a perceptual load search task, humanlike or anthropomorphic faces distracted participants in their task and in [51] the authors showed that a threatening humanoid robot, but not a social one, increased the level of participants' attention during the Stroop task. From these studies, it seems therefore that the sociality of the robot might constitute a distracting factor in diverse domains.

However, with respect to this hypothesis, our results seem to go in a different direction. We found that in adults the social interaction with a humanoid robot, perceived by participants as more humanlike, likeable, intelligent, and closer, did not affect human distraction, as suggested by the fact that the variability of responses (CV) does not increase, and the total error (RMSE) is even lower. Comparing the present study with related research on this issue, it is worth noticing at least three elements: the role of the robot, the cognitive load of the task and the demographics of participants. Specifically, in the present experiment, the perceptual task was designed to be intrinsically interactive so that the robot was not only present in the scene as a distractor [50], or a tutor/instructor [51], but it rather had the role of stimuli demonstrator for participants. This could explain the reason why the robot did not constitute a distraction for participants. Moreover, our results might also be explained by the fact that the reproduction task of this study was not cognitively or perceptually high demanding. Lastly, the experiment was designed to be performed by a demographic of adult participants. Context dependency has been already studied in visual perception of space in children [10], but only in the individual condition. Therefore, the question of whether a social robot distracts participants' perception is still open for this other age range.

From a comparison between the robot's behaviour in the social condition and the mechanical one, the robot's gaze seems to play a significant role. The social session was indeed designed to establish mutual gaze with participants between one trial and another and precede the hand moving towards the point predetermined for the touch. The ultimate purpose was to strengthen the belief of intentional behaviour in its human partners. On the contrary, when behaving mechanically, the robot directed its gaze in a static way toward a point diverged both from the participant and the touchscreen. As it has been viewed by [52], the behaviour of a robot responsive to their partners' gaze and establishing joint attention with them enhances both a favorable feeling of the users toward the robot and the users' belief of a favorable feeling of the robot toward them, if this behaviour is also supported by the eye-contact reaction. Also [53] showed the positive impact of eye-contact on human engagement in the interaction with the robot. In this sense, our results concur with these findings. The data gathered with the questionnaires highlight a substantial explicit preference for the social robot. Moreover, they are also supported by the behavioral measures of participants' gaze. The social robot's face was indeed looked more often both during and between trials revealing that the eye contact established by the robot after showing the stimulus was reciprocated and created a social context that was appreciated by participants.

As it has been explained in Materials and Methods, we opted for explicitly priming participants about the robot's intentionality and social skills. Our aim was to assess the phenomenon of context dependency in a social context in comparison with a non-social situation. We then attempted to reduce the variability of participants' beliefs about the meaning of the robot's "mechanical" behavior. Given the humanoid child-like shape of the robot, in fact, we could not exclude that some participants automatically would have anthropomorphized the robot, also interpreting the mechanical behavior not as such, but rather as social and negative (unfriendly or apathetic). We tried to minimize this risk with a design that foresaw congruent explicit and implicit information about the social (or non-social) nature of the interaction. It is relevant to note that the combination of explicit priming with the implicit behavioral cues produced a significant difference in participants impressions of the robot between the two sessions, but this difference was still very variable among participants. Moreover, the greater the difference in perception of the robot's anthropomorphism between the two social and mechanical conditions, the greater was the variation in context dependency at the perceptual level.

In general, it has been demonstrated that a robot can influence human attention [54], actions [55] and cognitive mechanisms [56], only by implicit behavioral cues. Considering these findings, we may hypothesize that robot's behavior might alone impact perception as well, in particular modulating context dependency. However, the present work does not allow to quantify the relative impact of robot's implicit cues and explicit priming. Now that the phenomenon has been proved, it will be interesting to verify in future studies whether either the robot's behavior or explicit priming along could impact participants' perceptions.

*C. Sharing perception*

The increased anthropomorphism and social intelligence attributed by participants and induced by all robot's behaviors seem thus to be the cause of a change in the perceptual schemes of the human interactant. In both conditions, the robot provides the stimulus to the human with the same biological movement of the arm. Still, only in the social session the perception ceases to be merely private for the human and becomes a perception of something shared with another agent: a shared perception. The context of shared perception influenced the entire perceptual process so that the integration of priors with sensory information was modified in favor of a major influence of the latter. It seems like the human observers were prone to evaluate more what was currently happening. Therefore, in shared perception, what is weighted more would be the shared source of information of the perceptual process, i.e. the current stimulus, rather than the private internal model, i.e. the prior. Perception becomes shared when another perceiving agent is present to our perception, something that in our experiment could happen rather with the social robot than with the mechanical one. This seems to produce a change in our perceptual mechanisms in that others' presence or behaviour affect the entire inferential process of perception from which our percept emerges.

Perception of something that is shared among two agents may therefore become shared itself. However, the multiple meanings related to the concept of "shared" requires clarification. What is shared among two agents is the real object to which perception refers and can be "shared" at least in two senses. In the first sense, two interactants can perceive the same – *shared* – stimulus coming from the environment. In this case, the real object of perception is shared because both perceive it simultaneously. Accordingly, perception becomes *shared* because the social context affects the way one agent perceives that thing. In the second sense, one can perceive what the other agent shows, i.e. *shared* by the other. In this case, the object of perception of one agent is what is shared by the other through an action, a bodily reaction or expression. Therefore, the observer's perception can be called *shared perception* because the observer, while perceiving something in the environment, incorporates into her/his perception the other's relation to that thing. In our study, both interactants were looking at the screen together. Also, it was the robot that provided the stimuli by showing the points on the touchscreen. It was, therefore, a shared perception in both senses. In the first sense of "shared", this means what the participants were aware to see together with the robot, whereas in the second sense, "shared" means rather what the robot was showing in each trial. It is true that also the mechanical robot showed stimuli to participants. However, its action was not unified to any apparent perception since it behaved as a mechanical arm: it was not a perceiving agent. That is the reason why in this case it is not possible to talk about shared perception. Only the social robot, thus, established a shared perception, a particular relation between the two social agents which significantly affected the observers' perception, as the results of this study demonstrate.

The idea of a shared perception raises how critical the ability of self-other distinction might also be in perception. Self-other distinction refers to the ability to distinguish others' representations from ours and is a key mechanism in empathy and, more generally, in understanding others [57], [58]. In this sense, the awareness that one's perception differs from others makes humans adopt peculiar perceptual mechanisms associated with social interaction. The enhanced reliance on the current stimulus associated with the variation of context dependency and induced by shared perception might be one of these social mechanisms of perception. Moreover, given the connection between the use of prior knowledge and developmental disorders proposed by [59], [60], the study of how sociality affects prior integration into perception becomes even more compelling.

*D. Future directions*

Therefore, the paradigm of shared perception and the study of inferential mechanisms of prior integration may bring a double outgrowth. On the one hand, it can be helpful to investigate more thoroughly and indirectly the self-other distinction mechanisms and how humans understand and align to others by focusing on the entire process of perception. On the other hand, shared perception becomes crucial also in human-robot interaction: it is a means to explore to what extent humans are affected by robots and interact with them as social partners. Further, it may promote the development of interactive machines designed to adapt to human abilities, and, therefore, enhance the outcome of the interaction. In several human-robot interaction scenarios, it would be indeed desirable to improve the quality of the interaction by reducing human perceptual errors caused by distraction, false prediction, the uncertainty of the sensory information. That is the case of collaborative robots in industries as well as robots in rehabilitation contexts. Both settings where gestures repetition, distraction, and uncertainty due to an occluded visual perspective or a deficit in sensory receptors, may adversely affect human perception. Nevertheless, interaction with social robots used to give information in public places or help older people in clinics and domestic environments, or even more with robots employed in developmental contexts, would be deeply enhanced if their design and behaviour were conceived based on human social skills. So, to advance toward improved collaborations between humans and robots, it is still needed to deepen the human perceptual mechanisms and the way they work during interactions.

## ACKNOWLEDGMENT

The authors would like to thank Alexander Mois Aroyo for his help in programming the robot iCub.

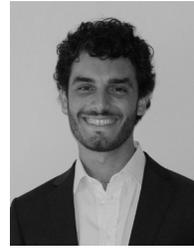
**Carlo Mazzola** (Graduate Student Member, IEEE) received his B.Sc. and, in 2018, his M.Sc. degree in Philosophy at the Università Cattolica del Sacro Cuore of Milan, Italy.

He worked as Visiting Researcher at the Italian Institute of Technology in the CONTACT Unit for the European Project wHiSPER. He is currently Ph.D. student at Bioengineering and Robotics at the University of Genoa, in collaboration with the department of Robotics Brain and Cognitive Sciences (RBCS) at the Italian Institute of Technology.

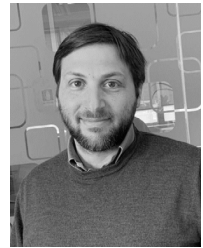
**Francesco Rea** (Member, IEEE) graduated in B.Sc. Information Engineering at the Università di Bergamo in 2004 and specialized in Computer Engineering at the Università di Bergamo in 2007. He got a M.Sc. degree in Robotics and Automation at the University of Salford, Greater Manchester University UK in 2008 and a Ph.D. degree in Robotics at the University of Genoa in 2012 contributing to different EU project (POETICON, eMorph).

He spent research periods at the Applied Cognitive Neuroscience laboratory of University of Lethbridge (Alberta, Canada) and at the Emergent Robotics Lab. of the Osaka University. He is currently Researcher at the Italian Institute of Technology (COgNiTive Architecture for Collaborative Technologies Unit), research leader of the Cognitive Interaction lab, and responsible for IIT of three European projects on collaborative robotics: H2020 APRIL, H2020 VOJEXT and HBP PROMEN-AID.

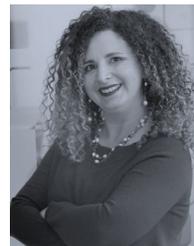
**Alessandra Sciutti** (Member, IEEE) received her B.Sc. and M.Sc. degrees in Bioengineering and the Ph.D. in Humanoid Technologies from the University of Genova in 2010.

After two research periods in USA and Japan, in 2018 she has been awarded the ERC Starting Grant wHiSPER (www.whisperproject.eu), focused on the investigation of joint perception between humans and robots. She published more than 80 papers and abstracts in international journals and conferences and participated in the coordination of the CODEFROR European IRSES project (https://www.codefror.eu/). She is currently Tenure Track Researcher, head of the CONTACT (COgNiTive Architecture for Collaborative Technologies) Unit of the Italian Institute of Technology (IIT) in Genova, Italy.

Dr. Sciutti is Associate Editor for several journals, among which the International Journal of Social Robotics, the IEEE Transactions on Cognitive and Developmental Systems and Cognitive System Research.